\documentclass[final]{cvpr}

\usepackage{times}
\usepackage{epsfig}
\usepackage{graphicx}
\usepackage{amsmath}
\usepackage{amssymb}

\usepackage[pagebackref=true,breaklinks=true,colorlinks,bookmarks=false]{hyperref}

\def\cvprPaperID{08} 
\def\confYear{CVPR 2021}

\begin{document}

\title{ I Only Have Eyes for You: The Impact of Masks On Convolutional-Based Facial Expression Recognition}

\author{Pablo Barros, Alessandra Sciutti\\
CONTACT Unit, Istituto Italiano di Tecnologia\\
Genova, Italy\\
{\tt\small {pablo.alvesdebarros, alessandra.sciutti}@iit.it}

}

\maketitle

\begin{abstract}

The current COVID-19 pandemic has shown us that we are still facing unpredictable challenges in our society. The necessary constrain on social interactions affected heavily how we envision and prepare the future of social robots and artificial agents in general. Adapting current affective perception models towards constrained perception based on the hard separation between facial perception and affective understanding would help us to provide robust systems. In this paper, we perform an in-depth analysis of how recognizing affect from persons with masks differs from general facial expression perception. We evaluate how the recently proposed FaceChannel adapts towards recognizing facial expressions from persons with masks. In Our analysis, we evaluate different training and fine-tuning schemes to understand better the impact of masked facial expressions. We also perform specific feature-level visualization to demonstrate how the inherent capabilities of the FaceChannel to learn and combine facial features change when in a constrained social interaction scenario. 
\end{abstract}

\section{Introduction}

The outstanding incursion of machine learning in our daily life in the last years changed drastically the manner we perceive, consume, and interact with artificial systems \cite{lecun2015deep}. Promising research in deep neural networks was quickly transformed into a heavy investment in the automation of image understanding, and one of the areas which were most affected is the perception of human faces \cite{kumari2015facial, anil2016literature}. Nowadays, we can find simple applications deployed in our smartphones that can detect, recognize, and even alter our faces \cite{katsis2015emotion, seng2019multimodal}. One of the areas where these impressive developments are still far behind, however, is on the processing, categorization, and understanding of emotions from facial expressions, in particular when addressed in scenarios where social interaction is somehow constrained \cite{kotsia2008analysis, roberson2012shades, carbon2020wearing}. In particular, in the current times of the COVID-19 pandemic situation, the application of affective perception on health care, assistive scenarios, and even tutoring becomes more important. In the current situation where masks are mandatory in most human interactions, persons do not only change how they express themselves, but also adapt how they perceive emotions from others. This is an important capability that is under-explored by the most successful affective perception solutions.

Most of the current state-of-the-art solutions for automatic facial expression recognition, however, claim to have addressed the problem by approaching maximum generalization \cite{ng2015deep, barros2017emotion, li2020facial}. The vast majority of these solutions deploy the computational power of artificial neural networks, boosted by data-driven deep learning of faces. The modus operandi of these solutions is to use millions of examples to tune these networks to extract specific facial features that, in controlled scenarios, represent and categorize affect. In most of these models, the learned features are comparable with existing human-made modelings such as the Facial Action Units\cite{ekman1997face}, which, together with a good performance in specific cases, puts them one step closer to their understanding and acceptance as a general facial representation system.

The problem these models face when deployed in socially constrained scenarios comes when combining these representations into affective categories \cite{li2020deep}. Most of these models, mostly for commodity and data availability, categorize affect using standard representations, whether a strict set of categories or dimensional pleasure/arousal/dominance scales. They are, thus, not only sensitive to representing only the facial features that are present on the training data but also on categorizing such features based on given affective labels. Such labels are usually obtained based on unconstrained scenarios, to maximize their capability of achieving a possible generalization of affective perception. 

When deployed in scenarios different from the ones they were tuned for, most of these models present difficulty to perform and even to adapt, given that deep neural networks are known to be extreme resources and data-hungry \cite{zhang2017facial}. These models are, thus,  extremely biased towards their application, and most of the time are difficult to adapt to specific scenarios \cite{tan2018survey}. In particular, the ones which do not have a popular interest, and thus, do not provide large amounts of available or labeled data. One of these scenarios, now in strong evidence given the COVID-19 pandemic, is when social interactions are constrained by the use of personal protective equipment such as facial masks. As most of these neural networks learn how to recognize affect based on a collection of facial features, if some of these features are not present, which is the case when using a mask (illustrated by Figure 1) these models tend to fail \cite{aly2019facial}. This effect can be observed, albeit on a smaller scale, in humans as well. However, due to our capability of changing the way we express emotions when using the mask \cite{wegrzyn2017mapping}, and consequently how we recognize expressions from persons with masks, we learn to compensate much better than any deep learning system.

Recently, we proposed the FaceChannel \cite{barros2020facechannel}, as a small and easy to adapt neural network, and evaluated it on different scenarios, including different fine-tuning methods. Our experiments showed that the network could be easily adapted due to its small number of parameters while maintaining good performance on most of the facial expression datasets. None of the evaluated datasets, however, encapsulated any constrained expression scenario, like the ones faced when recognizing facial expressions from persons with masks.

In this paper, we investigate how the FaceChannel \cite{barros2020facechannel} can adapt towards a constrained facial expression recognition scenario, where persons wear masks. For that, we leverage the one million images from the AffectNet dataset \cite{mollahosseini2017affectnet} by artificially including masks on them. Our evaluations include the investigation of different fine-tuning mechanisms on the FaceChannel and their impact on the network's final performance. Also, we investigate the differences between the FaceChannel trained with the normal AffectNet dataset and the MaskedAffectNet variation. We also provide an in-depth investigation on how the learned features emerge in both cases, and how they differ from each other, in particular in correlation with the performance of the models. Our analysis helps us to explain better the effect of constrained interaction scenarios on state-of-the-art convolutional-based facial expression recognition, and we discuss how different training schemes can address this problem.

\section{The Masked Face Channel}

The FaceChannel was recently formalized as a light-weighted convolutional neural network that is able to adapt towards different interaction scenarios. It showed to have a good adaptation towards novel scenarios \cite{barros2020facechannel, barros2020facechannels}, including the challenging task of recognizing affect during a Human-Robot Interaction setting \cite{belgiovine2020sensing}.

\begin{figure*}
    \centering
  \includegraphics[width=1.7\columnwidth]{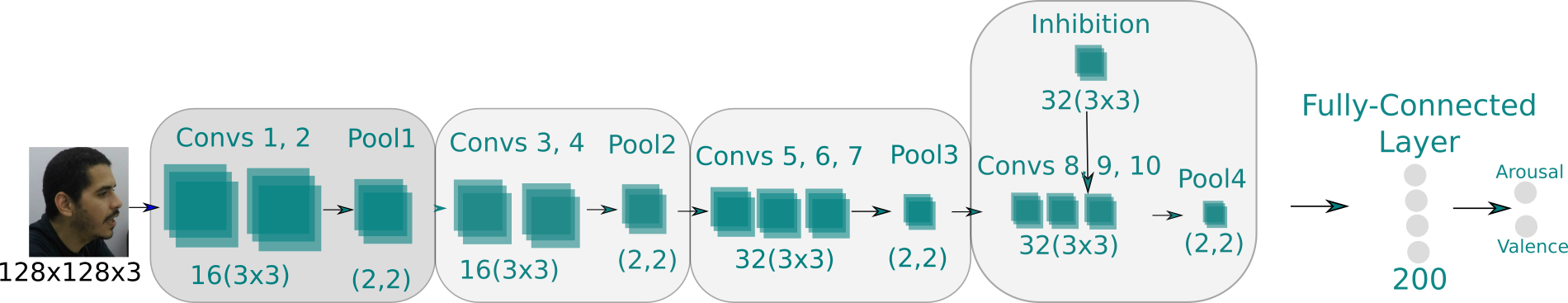}
  \caption{FaceChannel architecture used to learn facial expression representations and categorize them into a continuous arousal and valence values.}
\label{fig:faceChannel}
\end{figure*}

The FaceChannel has a total of 2 million parameters, which allows it to be trained from the scratch, or easily adaptable to other tasks without the need of an overwhelming computational power. In this paper, our implementation of the FaceChannel has $10$ convolutional layers, and applies  a shunting inhibitory layer ~\cite{Fregnac2003} in the last one. Each convolutional layer block is followed by $4$ pooling layers, as illustrated in Figure  \ref{fig:faceChannel}. The inhibitory layers were shown to improve adaptability towards facial expressions \cite{barros2016developing}, and an inhibitory neuron $S_{nc}^{xy}$, present at position ($x$,$y$) of the $n^{th}$ receptive field in the $c^{th}$ layer is defined as:

\begin{equation}
S_{nc}^{xy} = \frac{u_{nc}^{xy}}{a_{nc} + I_{nc}^{xy}}
\end{equation}

\noindent where $u_{nc}^{xy}$ is the activation function of the convolution unit, in our case ReLU,  and $I_{nc}^{xy}$ is the activation of the inhibitory units. The passive decay term $a_{nc}$ is also updated during training and is shared among each inhibitory filter. 

When training the FaceChannel for the AffectNet and MaskedAffectNet datasets, we implement a fully-connected ReLU-based hidden layer after the convolutional layers. This layer is followed by two output layers, each of them implementing a linear activation. Each of these layers allows a continuous and dimensional representation of affect through arousal and valence.

To optimize the FaceChannel towards the AffectNet and MaskedAffectNet, we use a parameter exploration based on a Three-Parzen Exploration method ~\cite{bergstra2011algorithms}. Table \ref{tab:searchSpaceAll} reports our search space and final selected parameters. We maximize the optimization based on the final accuracy of the model.

\begin{table}
\center
\begin{tabular}{ |c|c|} 
\hline
 \textbf{Parameter} & \textbf{Search Space} \\ \hline
 \multicolumn{2}{|c|}{\textbf{Fully-Connected Layer}}\\\hline
 Number of Layers & [\textbf{1},2,3]\\
 Units Per Layer & [16, 32, 64, 128, \textbf{256}, 512, 1024] \\
 Learning Rate & [0.0005; 0.9] - Selected: \textbf{0.05}\\
 Opitmizer & [ADAM, \textbf{SGD}]\\
 \hline
\end{tabular}
\caption{Search space and selected parameters used to optimize the FaceChannel.}
\label{tab:searchSpaceAll}
\end{table}

\section{Evaluation}

Our main goal is to investigate the differences that a masked facial expression cause on the FaceChannel feature representation and final performance. Thus, we rely on the variety of samples and a good label distribution present on the AffectNet dataset to guarantee a proper facial feature emergence on the FaceChannel. The MaskedAffectNet modifies the images of the AffectNet, but maintains the same label distribution, which guarantees a fair comparison base in our investigations.

\textbf{The AffectNet~\cite{mollahosseini2017affectnet}} is our main evaluation patform, and it has more than $1$ \textit{million} images, crawled from the internet, with half of them manually annotated using mechanical Turk. Each image has a single label based on a continuous arousal and valence value, ranging from [$-1$ to $1$]. The dataset authors separated it into specific training and validation subsets, which we use in our experiments. The main metric we use to measure perofrmance is the concordance correlation coeficient (CCC) \cite{Lawrence1989} between the models' outputs and the true labels of the images. The FaceChannel showed to be competitive with other convolutional-based models when evaluated on the AffectNet  \cite{barros2020facechannel}, and we use this results as our performance baseline.

\begin{figure}
    \centering
  \includegraphics[width=0.7\columnwidth]{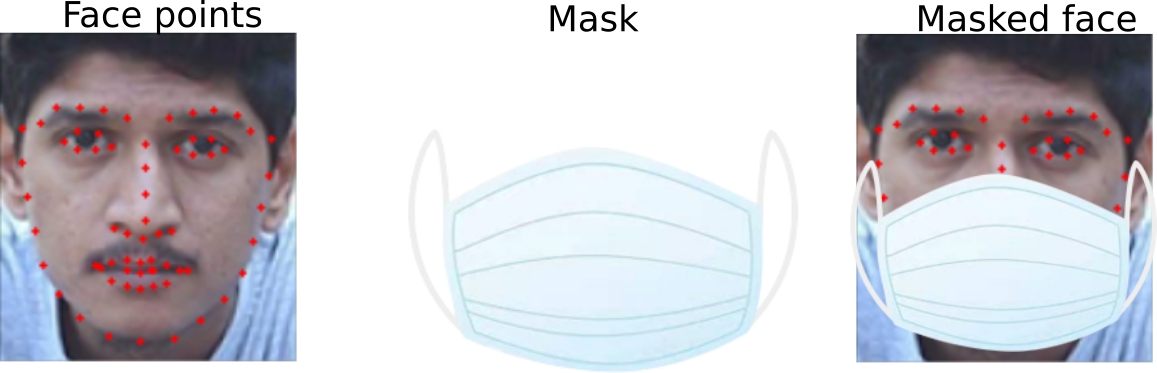}
  \caption{Example of the processing to add masks to the AffectNet: we find the facial keypoints on the images, and apply a geometrical transformation the mask to match the mouth key points.}
\label{fig:maskedFace}
\end{figure}

\textbf{The MaskedAffectNet} dataset is proposed here to represent a constrained interaction scenario. It is composed of the same images of the AffectNet dataset, but with the artificial addition of a facial mask. The mask is added in a postprocessing scheme that finds the facial points of the mouth, and using a geometrical transformation on a standard face mask image, fixes the mask on top of the mouth, as illustrated in Figure \ref{fig:maskedFace}. The results reassemble closely the use of the mask in a real-world environment, but following the same label scheme of the AffectNet.

\subsection{Experimental Setup}

We conduct our experiment in two settings: first, we perform a series of baseline studies to investigate the impact that the mask has on the FaceChannel. Second, we investigate the emergence of the facial features by calculating the saliency maps of the last convolutional channels of the model, when processing images from the AffectNet and the MaskedAffectNet.

For our first set, we run four experiments: first, we investigate the performance of the FaceChannel when trained on the AffectNet and evaluated on the MaskedAffectNet. Then, we run train the FaceChannel from the scratch on the MaskedAffectNet dataset and perform the same evaluation. Finally, we perform two fine-tuning routines: first investigating the re-training of the last channel of the FaceChannel, and the entire network. The entire experimental summary is reported in Table \ref{tab:experimentalSummary} Our goal with this experiment is to evaluate the network's performance in these different settings.

\begin{table}[h]
    \center\begin{tabular}{ |c c c|}
        \hline
        Trained On & Fine-Tunned On & Method       \\ \hline
        AffectNet           & -  & Scratch     \\ 
        AffectNet         & MaskedAffectNet  & Scratch   \\ 
        AffectNet         & MaskedAffectNet  & Last Conv    \\ 
        AffectNet         & MaskedAffectNet  & All Layers    \\ 
        MaskedAffectNet         & -  & Scratch        \\ \hline
    \end{tabular} 
    \caption{Experimental summary of our first baseline setting. For each of the experimental setting, we evaluate the networks' performance on both the AffectNet and MaskedAffectnet.}

\label{tab:experimentalSummary}
\end{table}

Second, we run an exploratory routine that calculates the saliency map of individual images from both the AffectNet and the MaskedAffectNet for all the models we evaluated in our first set. We discuss how each of the training and fine-tuning routines impacts the feature formation and the performance of the model.

\section{Results}

\begin{table*}
    \center\begin{tabular}{ |c c c c c c c |}
        \hline
        Trained On & Fine-Tunned On & Method & \multicolumn{2}{c}{AffectNet} &  \multicolumn{2}{c|}{MaskedAffectNet}        \\ \hline
        & & & Arousal & Valence & Arousal & Valence \\ \hline
        AffectNet           & -  & Scratch &  0.46 & 0.61 & 0.18 & 0.16  \\ 
        MaskedAffectNet         & -  & Scratch   &  0.25 &0.33 &0.43 &0.48  \\
        AffectNet         & MaskedAffectNet  & Last Conv &  0.34 & 0.39 &0.33 &0.43  \\ 
        AffectNet         & MaskedAffectNet  & All Layers  &  0.38 & 0.45 &0.45 &0.53  \\ 
         \hline
    \end{tabular} 
    \caption{Experimental summary of our first baseline setting. For each of the experimental setting, we evaluate the networks' performance on both the AffectNet and MaskedAffectnet. We report Arousal and Valence in terms of Concordance Correlation Coefficient (CCC).}

\label{tab:baselineExperiments}
\end{table*}

Our baseline experiments are fully reported in Table \ref{tab:baselineExperiments}. We observe that when the FaceChannel is trained with the AffectNet \cite{mollahosseini2017affectnet} and evaluated with the MaskedAffectNet, the performance drops considerably. When inverting the setting, and training the network with the MaskedAffectNet, and evaluating with the AffectNet, the performance does not drop much. This is an indication that the network trained with the MaskedAffectNet somehow learns a more robust, and thus shareable, facial representation.

The best results on the MaskedAffectNet dataset, with a total CCC of 0.53 for arousal and 0.45 for valence, were obtained when fully re-training the network, and are expected as we guide the entire facial representation towards the presence of masks. This, however, can be costly, as the number of parameters to be adapted is much higher than when re-training only the last convolutional layers. Also, re-training the entire network obtained a higher CCC when compared to re-training the entire network from the scratch, which is an indication that some of the features learned by the AffectNet-trained network somehow contribute to the MaskedAffectNet recognition. 

When evaluating the MaskedAffectNet-trained models with the AffectNet, we observe that while re-training the entire model yields the best performance, re-training the last convolutional layers achieves a close CCC. In this particular experiment, the facial features were learned with full face expressions from the AffectNet, which indicates that simple high-level feature recombination obtained when training the last convolutional layer is enough for this scenario.

\begin{figure}
    \centering
  \includegraphics[width=1\columnwidth]{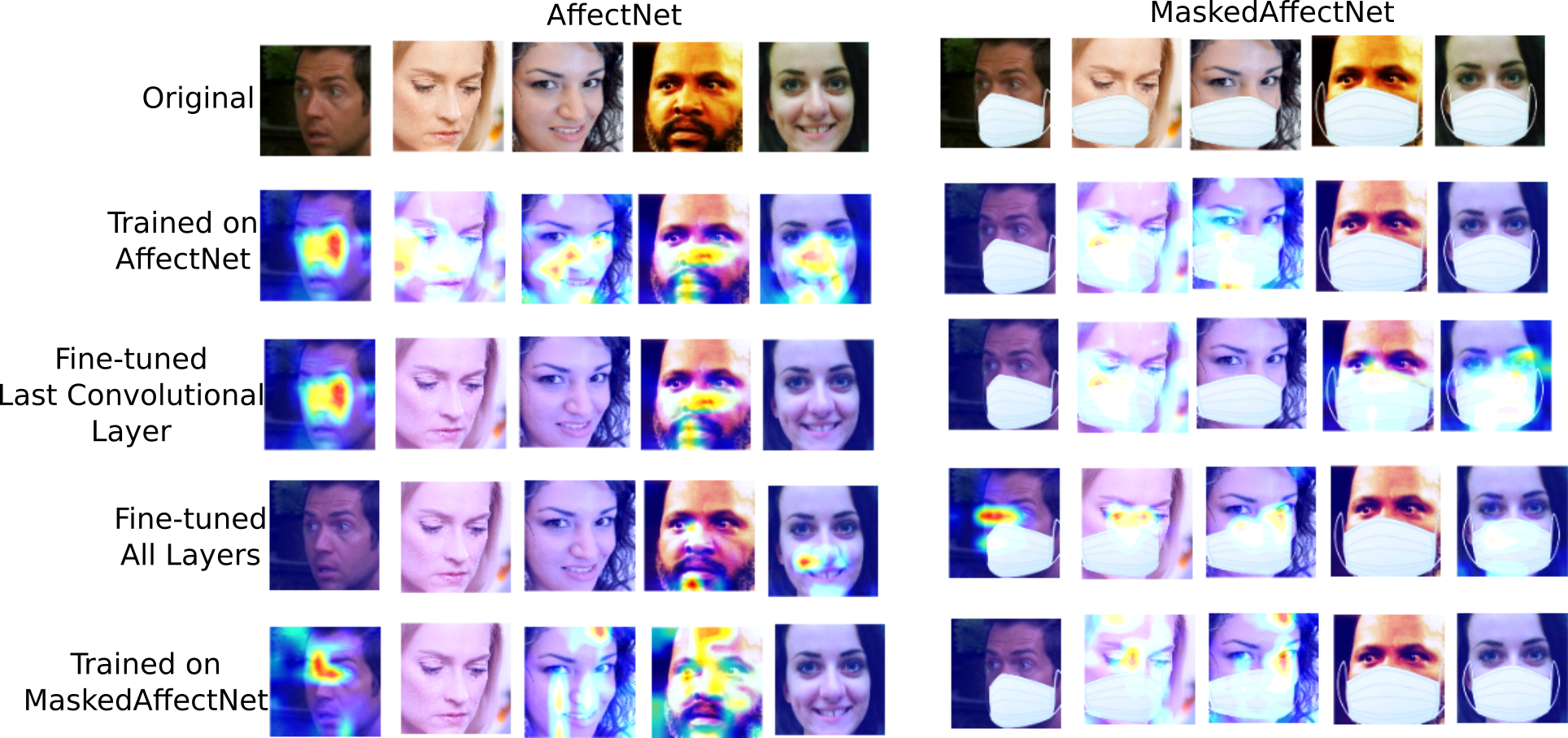}
  \caption{Saliency map visualization of images from the AffectNet and MaskedAffectNet datasets processed by all the evaluated models.}
\label{fig:visualization}
\end{figure}

Our second experimental setting involves visualizing the saliency maps of each of the trained models when processing images from the AffectNet and the MaskedAffectNet datasets. We plot these results in Figure \ref{fig:visualization}. We observe the differences from a model trained on the AffectNet and the MaskedAffectNet images: the focus on the mask. The performance drop of the AffectNet-based model is easily explained by the presence of facial representations from the mouth and chin, which are covered when the mask is present. We observe, as well, that the model which was trained only in the last convolutional layer somehow re-combine the representations to give more strength to the upper-face region features, including eyes and cheeks. We also observe that on the models trained from the scratch, and the fully fine-tuned model, the mask itself becomes a feature. These models identify that the size and position of the mask somehow depict some facial characteristics, which probably explain the performance boost on these training processes.

\section{Discussions}

In this paper, we present an in-depth study on how FaceChannel adapts towards recognizing facial expressions in very specific constrained settings: facial expressions from masks. Our experiments demonstrate that the network trained with the general AffectNet dataset struggles to deal with the partial expressions from the masked faces. Pre-training the network, however, increases the performance of the model, showing that it is possible to adapt it towards this scenario. Beyond the network's performance, it is important to notice also the impact of our analysis in two fronts: how partial expressions can be recognized and represented, and how important is to have a reliable adaptation of facial expression recognition models.

\subsection{Partial Expression Recognition}

Our analysis can be seen as a simulation of the real-world scenario. Although the AffectNet dataset has millions of images, artificially adding masks to it does not necessarily approximate it from real-world interactions. What we focus on, however, is on the capability of the FaceChannel to depict and to adapt to these changes, modifying its learned representations as needed. Our feature-level analysis shows that by re-training only the last convolutional layer of the FaceChannel, improving drastically its performance, we already can change the focus of the network's encoder towards the upper facial area. Although the mouth is extremely important for depicting facial expressions, as shown by the feature visualization of the original FaceChannel, the network trained with the masked data retains a good performance. This indicates two important things: first, at least for the images from the AffectNet dataset, most of the affective content can be depicted from the upper-face region with a still good performance. Second, the masked FaceChannel learned to integrate the mask on its feature representation. Although the lower face is covered, the presence of the mask following the jaw-line can be an indication of mouth open and close, for example. The network uses the mask shape to somehow represent these characteristics.

\subsection{The Impact of Feature Adaptation}

On a deeper look into the computational effort to re-train the FaceChannel, and the final results, we can relate the performance of the model with the meaning behind each training strategy. When re-training the entire model, we are changing drastically the feature representation of the network together with its own decision making, towards a very specific scenario. Although in our analysis the tasks are similar, actually even using the same dataset, the performance drop demonstrates how facial expression perception can be sensitive to simple image occlusion. Having a model that is easily adaptable, in particular when deployed in constrained computational environments such as mobile robots, is certainly an advantage.

\section{Conclusion}

In this paper, we present a study on the impact that facial masks have on facial expression recognition neural networks. We use the recently proposed FaceChannel \cite{barros2020facechannel}, to investigate how faces with masks change the performance of the model, and how different fine-tuning schemes can be used to mitigate the performance drop. We propose here the MaskedAffectNet dataset, a version of the AffectNet dataset that has images with facial masks but maintains the same affective arousal and valence-based labels, to help the direct comparison of our evaluation. By visualizing the saliency maps of processed images from both datasets, we can pinpoint that the network trained with the AffectNet relies heavily on the mouth and chin regions to make affective decisions, which contributes to poor performance on the masked faces. However, when re-adapting these features towards the upper-face, as a byproduct of training and fine-tuning the network with the MaskedAffectNet dataset, the performance on the original AffectNet is not much affected. Also, we show that the FaceChannel adapts towards masks by using the mask itself as a feature and probably analyses the size and position of the mask as an affective determinant characteristic.

Our analysis, although in-depth on the FaceChannel, needs to be extended towards other convolutional-based models, and other affective scenarios. As a clear future work direction, we will continue our investigation also taking into consideration different learning schemes, such as self-supervised representation learning, and interaction scenarios, including settings involving social robots. We intend to evolve the FaceChannel towards an adaptable neural network that can be easily deployed on different devices, relying on specific fine-tuning strategies involving attention layers. We are also very interested in viewing facial expression perception as a life-long adaptive goal, and thus, studying continual and open-ended learning strategies would be very beneficial.

\section*{Acknowledgement}
Thid sutudy is supported by a Starting Grant from the European Research Council (ERC) under the European Union's Horizon 2020 research and innovation programme. G.A. No 804388, wHiSPER.

{\small
\bibliographystyle{ieee_fullname}
\bibliography{egbib}
}

\end{document}